\documentclass[conference]{IEEEtran}
\IEEEoverridecommandlockouts

\usepackage{cite}
\usepackage{amsmath,amssymb,amsfonts}
\usepackage{algorithmic}
\usepackage{graphicx}
\usepackage{textcomp}
\usepackage{xcolor}

\usepackage{tikz}
 \usepackage{hyperref}
 \usepackage{algorithm}
\usepackage{algorithmic}

\def\BibTeX{{\rm B\kern-.05em{\sc i\kern-.025em b}\kern-.08em
    T\kern-.1667em\lower.7ex\hbox{E}\kern-.125emX}}
\begin{document}

\title{Taming the Exponential: A Fast Softmax Surrogate for Integer-Native Edge Inference
}

\author{
  \IEEEauthorblockN{{
    Dimitrios Danopoulos\textsuperscript{1},
    Enrico Lupi\textsuperscript{1},
  }}

  \IEEEauthorblockN{{
    Michael Kagan\textsuperscript{4},
    Maurizio Pierini\textsuperscript{1}
  }}

  \IEEEauthorblockA{
    {\textsuperscript{1}European Organization for Nuclear Research (CERN), Geneva, Switzerland}\\
    {\{dimitrios.danopoulos,enrico.lupi,maurizio.pierini\}@cern.ch}
  }

  \IEEEauthorblockA{
    {\textsuperscript{4}SLAC National Accelerator Laboratory, Menlo Park, CA, USA}\\
    {makagan@slac.stanford.edu}
  }
}

\maketitle

\begin{abstract}
Softmax can become a computational bottleneck in the Transformer model's Multi-Head Attention (MHA) block, particularly in small models under low-precision inference, where exponentiation and normalization incur significant overhead. As such, we suggest using Head-Calibrated Clipped-Linear Softmax (HCCS), a bounded, monotone surrogate to the exponential softmax function, which uses a clipped linear mapping of the max centered attention logits. This approximation produces a stable probability distribution, maintains the ordering of the original logits and has non-negative values. HCCS differs from previous softmax surrogates as it includes a set of lightweight calibration parameters that are optimized offline based on a representative dataset and calibrated for each individual attention head to preserve the statistical properties of the individual heads. We describe a hardware-motivated implementation of HCCS for high-throughput scenarios targeting the AMD Versal AI Engines. The current reference implementations from AMD for this platform rely upon either bfloat16 arithmetic or LUTs to perform the exponential operation, which might limit the throughput of the platform and fail to utilize the high-throughput integer vector processing units of the AI Engine. In contrast, HCCS provides a natural mapping to the AI Engines' int8 multiply accumulate (MAC) units. To the best of our knowledge, this is the first int8 optimized softmax surrogate for AMD AI engines that significantly exceeds the speed performance of other reference implementations while maintaining competitive task accuracy on small or heavily quantized MHA workloads after quantization-aware retraining.
\end{abstract}


\section{Introduction}
Transformers are based on Multi-Head Attention (MHA). For MHA, softmax is used to transform dot-products into probability distributions. The matrix multiplication of attention has been heavily optimized in recent years on modern hardware platforms; however, the softmax can still represent a performance and efficiency bottleneck in certain cases due to its characteristics of reduction, exponentiation and normalization. In small and edge-deployed Transformer models targeted for low-precision inference, this overhead is much greater since the ratio of non-GEMM operations to GEMM operations grows as the size of matrices decrease. Although the attention logits are quantized down to low bit-widths to improve efficiency (for example, int8), the softmax function is generally computed using either look-up table (LUT) or floating-point arithmetic \cite{10980492}. In practice, this generally results in converting the quantized logit to a floating-point value and performing an expensive exponential operation. On many accelerators and for some workloads (particularly those in cloud-based environments), the overhead of these two operations (i.e., precision crossing and the associated overhead) may be tolerable.

AI Engines from AMD are a type of domain-specific accelerator integrated into the Versal adaptive compute platform. These are intended to provide high-throughput and low-latency solutions for machine-learning workloads at the edge \cite{aie4ml_paper}. Each AI Engine core includes wide-vector units optimized for integer arithmetic, especially for int8 and bfloat16 (BF16) multiply—accumulate (MAC) operations. There is also software-programmable local memory that accompanies each core. AMD provides a reference softmax implementation in BF16 which uses explicit exponential computations. Even though the exponential operation can be performed in BF16, it is still very expensive and represents a significant source of latency. On AI Engine (AIE-ML and AIE-MLv2), LUT accesses occur in 16-bit granularity and exhibit limited throughput. Therefore, softmax implementations are either compute-bound by the exponential approximations or memory-bound by LUT fetches (i.e., up to 4 in parallel in AIE-ML). As a result, there is no native int8 softmax implementation currently available. The use of floating-point softmax in int8-quantized models requires the additional conversions from int8 to float and float to int8. These conversions involve additional unpacking, casting, and pipeline stages that consume a significant portion of the execution cycles. Therefore, softmax implementations in quantized models cannot seamlessly achieve the same throughput as the native int8 computation pipeline on AI Engine.

For this reason, the desire for a different approach to softmax for quantized attention was motivated by the previous observations. From an algorithmic standpoint, attention softmax has a constrained operating regime: logits are max-centered, extreme negative values contribute negligibly to the final distribution, and only the relative ordering and approximate ratios of the probabilities are important. This implies that exact exponentiation is not always required. From a hardware standpoint, it is preferable to avoid both the floating-point conversion overheads and the LUT-dominant designs. Instead, it is preferable to implement softmax in terms of operations that are consistent with the native execution units of the accelerator. In this work, we introduce \emph{Head-Calibrated Clipped-Linear Softmax (HCCS)}, a monotonic and bounded softmax surrogate designed for quantized multi-head attention. It preserves ordering and generates numerically stable probability distributions without the need for explicit exponentiation. Furthermore, we introduce a lightweight per-head calibration method that adjusts surrogate parameters offline using representative data, enabling accurate approximation across heterogeneous attention head distributions. This reformulation allows softmax computation to be mapped directly onto the native int8 MAC pipeline on AMD Versal AI Engine, avoiding expensive exponential operations and memory-bound LUT accesses.

The main contributions of this work are:
\begin{itemize}
  \item We propose \emph{Head-Calibrated Clipped-Linear Softmax (HCCS)}, a monotonic and bounded softmax surrogate for quantized multi-head attention that preserves ordering and produces numerically stable probability distributions without explicit exponentiation.
  \item We introduce a lightweight per-head calibration method that adjusts surrogate parameters offline using representative data, which allows HCCS to generate accurate approximations over a variety of heterogeneous attention head distributions.
  \item We describe the first int8-optimized softmax proxy implementation on AMD Versal AI Engine. We map the softmax computation directly onto the native int8 MAC pipeline. 
  \item We show that HCCS achieves significantly higher throughput than AMD's BF16 reference softmax on AI Engine while maintaining the stability of attention behavior on small and quantization-stressed MHA workloads.
\end{itemize}

\section{Related Work}
The increasing importance of softmax and its variations comes from how attention uses row-wise normalization with expensive non-linearity operations and global reduction operations (sum, max, reciprocal) inside every transformer layer. 
Once the surrounding matrix multiplications have been quantized and optimized, the remaining (de)quantize $\rightarrow$ Softmax $\rightarrow$ (re)quantize" sequence could be the most important on edge devices and would motivate integer- and hardware-friendly versions of softmax.

\subsection{Integer-domain exponential and softmax approximations}
There has been considerable research in preserving the softmax structure and replacing the exponential operation with a cheaper version in low-precision. I-BERT \cite{kim2021ibert} provides integer-only transformers for all layers and includes integer approximations for all non-linearities. For softmax specifically, I-BERT applies max-subtraction for stability and splits the argument to the exponential function into a logarithmic quotient and remainder such that exp(x) is equal to a product of a bound-range exponential and a power-of-2 term that can be implemented using integer left/right shifts. The bound-range portion is then approximated as a low order polynomial. More recent integer attention pipeline designs build upon the idea of incorporating approximation into both normalization and quantization. IntAttention \cite{zhong2025intattention} builds upon this by including the IndexSoftmax approximation method, which (i) performs max-subtraction and sparsity aware clipping in the integer domain, (ii) approximates the exponential using a compact 32 entry look-up table (LUT), and (iii) performs row-wise integer normalization directly into an 8-bit probability tensor without ever using floating point. In terms of accelerators, ITA \cite{islamoglu2023ita} presents a streaming integer softmax designed to handle quantized attention on embedded hardware. Its design emphasizes on-the-fly processing to minimize the number of memory passes and limits the precision of intermediate values used during the inversion and accumulation process, allowing for a small, energy efficient softmax module to be integrated into a larger attention accelerator.

\subsection{Hardware--software co-design for synchronization and reduction bottlenecks}
While these designs approximate the exponential function, other works focus on reducing the hardware overhead associated with the reduction and synchronization steps required for softmax.

Softermax \cite{stevens2021softermax} replaces $e^{x}$ with $2^{x}$ to allow for shift friendly renormalization and fuses the max computation into an online normalization step. It removes the need for a separate reduction, providing better energy/area efficiency than previous implementations of softmax in accelerator designs. ConSmax \cite{10.1145/3676536.3676766} reduces the need for both max search and denominator summation at inference time by utilizing learnable normalization parameters. This approach sacrifices exact unit sum probabilities in favor of increased parallelism and a smaller dedicated hardware unit (for example, bit-width split LUT structures and simple scaling data paths).

\subsection{Softmax replacements}
Another approach eliminates the need for softmax altogether. Sparsemax \cite{martins2016sparsemax} projects the input logits onto the simplex via Euclidean projection. This results in interpretable and sparse output probabilities. However, it also requires the introduction of additional primitives for sorting/selecting and thresholding (such as a naive $O(K\log K)$ evaluation via sorting). These primitives can be less amenable to hardware implementation compared to purely arithmetic pipelines. Rectified Linear Attention (ReLA) replaces softmax with a ReLU based transformation. It reports competitive translation quality when appropriate stabilization is applied~\cite{rela2021}. Similarly, alpha entmax and its adaptive variants generate controllable sparsity and exact zeros in attention weights by generalizing softmax to a class of sparse transformations~\cite{entmax2019}.

\subsection{HW implementations in practice}
On current GPUs, softmax can become a significant bottleneck because exponentiation is often executed through FP32 pathways and incurs datatype-conversion overhead. TurboAttention \cite{kang2024turboattention} introduces the Sparse Activated Softmax (SAS). SAS combines a hybrid LUT-polynomial exponential approximation with a negligible-exponential pruning and integrates it into the FlashAttention style kernels. To avoid the use of the expensive transcendental units, softmax is usually implemented in the FPGA tool flows using the LUT based exponentials and reciprocal approximations. For instance, an implementation of the Transformer in the hls4ml-based Transformer \cite{jiang2024hls4mltransformer} transforms the softmax computation to $\{e^{z_i}\}$ once, reduces once, inverts once and multiplies. This is represented as a pipelined Exp-LUT/reduction/Inv-LUT/scaling datapath. However, the hls4ml framework \cite{schulte2025hls4ml} discuss LUT based activations and compilation time table generation across quantized precisions to compute an activation function of whatever complexity. In case of parallel accesses, BRAM duplication may be required to ensure single clock access. Finally, on AI-engine class accelerators, AMD provides reference implementations that emphasize numerical stability. The Vitis AI Engine softmax tutorial \cite{amd_softmax_vitis} describes bfloat16 softmax kernels that avoid overflows via max-subtraction and accelerate exponentials via IEEE-754 exponent bit construction. It reports cycle level performance for large-class softmax vectors. Similarly, AMD’s open-source IRON project for Ryzen AI NPUs includes a bfloat16 softmax operator in its operator dashboard, indicating support for such kernels in the close-to-metal NPU software stack \cite{11008991}. 

\subsection{Positioning relative to our HCCS clipped-linear surrogate}
Our HCCS differs from all other exponential approximation methods (e.g. \cite{kim2021ibert,zhong2025intattention,kang2024turboattention}) by completely replacing the exponential mapping with a calibrated clipped linear surrogate. Thus, we do not need to store any LUT or perform any polynomial MAC chain. Unlike Sparsemax \cite{martins2016sparsemax}, we also do not have to sort/select primitives. Like synchronization free alternatives such as ConSmax \cite{10.1145/3676536.3676766}, we can preserve unit sum normalization while retaining the row wise reduction but guaranteeing a valid probability simplex. However, this is at the cost of one synchronization barrier. Overall, HCCS targets minimal primitive integer datapaths (adds, subtracts, compares, shifts/multiples, and a reciprocal/normalization), with calibration (e.g. head-wise) being applied to match empirical attention distributions.

\section{Implementation of a Fast Softmax Surrogate}
\label{sec:hccs_impl}

The inference-time implementation of the proposed Head-Calibrated Clipped-Linear Softmax (HCCS) surrogate and the design decisions made to achieve a high-throughput implementation on AMD Versal AI Engines are discussed in this section. The primary goal was to replace the computationally expensive exponential and floating-point normalization normally found in the softmax function with a few integer operations (max-reduction, subtraction, clamping, multiply—accumulate, and reciprocal-based normalization) that can be efficiently mapped to the AIE int8 vector MAC pipeline.

\subsection{Integer-Form Surrogate and Range Mapping}
\label{subsec:int_form}

Given quantized attention logits $\mathbf{x}\in\mathbb{Z}_{8}^{n}$ (per row), standard max-centering computes
\begin{equation}
d_i = x_i - \max_j x_j \le 0.
\end{equation}

Directly implementing $d_i$ as a signed value would obviously produce negative values and necessitate wider signed arithmetic for all subsequent elements of the pipeline. Many hardware architectures have a pure int8 representation that has the most efficient vector subtract and clamp pattern for this workload:

\begin{equation}
\delta_i = \min\!\big(\max_j x_j - x_i,\; D_{\max,h}\big),
\qquad \delta_i \in [0,\, D_{\max,h}],
\label{eq:uint_distance}
\end{equation}

where $D_{\max,h}$ is a per-head clamp bound. Algebraically, the above transformation is equivalent to max-centering followed by negation and clamping. However, it ensures that the intermediate values fall into a compact interval which may be represented using \texttt{uint8}. The clamping prevents extreme outlier values (large negative $d_i$) from dominating the fixed-point dynamic range, while at the same time preserving the relative ordering of the input logits within the active window. Using $\delta_i$, the linear surrogate is written as a \emph{decreasing} function of distance:
\begin{equation}
s_i = B_h - S_h\,\delta_i,
\label{eq:linear_distance_form}
\end{equation}
with $B_h > 0$ and $S_h \ge 0$. This form is attractive on many
hardware accelerators because it is a single int8 MAC followed by
normalization. Non-negativity of $s_i$ for all
$\delta_i \in [0, D_{\max,h}]$ is guaranteed by enforcing the
calibration constraint
\begin{equation}
B_h - S_h\,D_{\max,h} \ge 0,
\label{eq:nonneg_constraint}
\end{equation}

which makes an explicit per-lane $\max(0,\cdot)$ rectifier redundant
(see also Section~\ref{subsec:hw_mods}).

\subsection{Normalization in Fixed-Point}
\label{subsec:normalization_fixed_point}

The surrogate scores are normalized to form a probability distribution:
\begin{equation}
Z = \sum_i s_i, \qquad p_i = \frac{s_i}{Z}.
\end{equation}

The entire normalization process was designed to be executed completely in integer arithmetic and will not require any intermediate conversions to/from floating-point representations. The accumulated sum $Z$ is calculated with 32-bit precision to avoid potential overflow. A reciprocal scaling factor $\rho$ is also calculated once per row:

\begin{equation}
\rho = \left\lfloor \frac{T}{Z} \right\rfloor,
\label{eq:rho_def}
\end{equation}
where $T$ is the target integer scale ($T = 32767$ for int16 output or $T = 255$ for int8 output). The value $\rho$ fits within 16 bits and is broadcast directly into the vector lanes. The resulting outputs
\begin{equation}
\hat p_i = s_i \cdot \rho
\end{equation}

represent scaled probabilities such that $\hat p_i \in [0,T]$ and $\sum_i \hat p_i \approx T$ up to integer truncation error, allowing the probabilities to remain in integer form while preserving compatibility with fixed-point attention pipelines.

\paragraph{Rationale for Q0 Reciprocal Design.}
Although a Q15 reciprocal has additional fractional precision, when broadcast into int16 vector lanes, there will be intermediate products that will exceed the int16 range, which means that one would need to do either 32-bit vector multiplications, or add another down-sampling stage. Therefore, we use a Q0 reciprocal formulation~\eqref{eq:rho_def} that has enough precision for attention normalization, and preserves full compatibility with the native and smaller int16 vector pipeline.

\paragraph{int8 output path.}
If the desired output format is int8 (i.e.\ $T=255$), the normalization will use a shifted fixed-point reciprocal to retain more precision before the final down-shifting. More precisely, the kernel calculates:

\begin{equation}
\rho_{\mathrm{u8}} = \left\lfloor \frac{255 \cdot 2^{R}}{Z}
\right\rfloor,
\label{eq:rho_u8}
\end{equation}

where $R$ is a platform-specific right-shift constant (\texttt{INV\_SHIFT}$=15$ in the reference implementation). The intermediate product $s_i \cdot \rho_{\mathrm{u8}}$ is then right-shifted by $R + \texttt{OUT\_SHIFT}$ bits to get the final uint8 values. This approach keeps the fractional precision in the multiplication phase, and thus avoids 32-bit vector lanes during the normalization phase.

\paragraph{Reciprocal Approximation via Leading-Bit Detection.}
To further reduce latency, the reciprocal $\rho$ may optionally be approximated using leading-bit detection (CLB). Let
$k = \lfloor \log_2 Z \rfloor$. Then
\begin{equation}
\rho \approx \frac{T}{2^k}.
\end{equation}

Since $2^k \le Z < 2^{k+1}$, this approximation overestimates the ideal reciprocal by at most a factor of two, replacing the scalar divide with a bit-shift derived from the position of the most significant set bit of $Z$. Since $Z$ is usually far from being a power of two, the overestimate is typically much less than a factor of 2. In our measurements, the speedup achieved through this substitution is greater than  $3\times$  for short sequences, where the latency of calculating the reciprocal is not averaged out over a long sequence.

\subsection{Offline Calibration}
\label{subsec:calibration_and_frozen}

There are some small number of per-head constants $\theta_h = (B_h, S_h, D_{\max,h})$, that are determined offline and held fixed during inference. Holding these parameters fixed defines a stable, hardware-constrained attention function to which the rest of the model parameters are adjusted during quantization-aware training. This is analogous to holding the quantization bounds fixed during quantization-aware training: the nonlinearity is fixed, but the network adapts to compensate for its own errors. From a deployment viewpoint, the fixed parameters allow us to enforce explicitly the necessary constraints of monotonicity, bounded range, and non-negative values, particularly $B_h - S_h D_{\max,h} \ge 0$ and $D_{\max,h} \le 127$, which are essential for a correct and overflow-free int8 datapath. There is a learnable version of HCCS in principle, e.g. by treating $\theta_h$ as differentiable parameters under constrained optimization. We view this as complementary to the current work, and defer consideration of this case until future work. Given the fixed-parameters setting, the calibration problem can be written as:

\begin{equation}
\begin{split}
(\hat{B}_h, \hat{S}_h, \hat{D}_{\max,h}) &=
  \arg\min_{\,B,\,S,\,D}\;
  \mathbb{E}_{\mathbf{x}\sim\mathcal{D}_h}\!
  \Big[ \mathrm{KL}\big(\mathrm{softmax}(\mathbf{x}) \\
  &\quad\|\; \hat{p}^{\mathrm{HCCS}}(\mathbf{x};\,B,S,D)\big) \Big]
\end{split}
\label{eq:calib_objective}
\end{equation}

subject to the integer deployment constraints of Section~\ref{subsec:int_range_constraints}. Here $\mathcal{D}_h$ represents the empirical distribution of the int8 attention logits at head $h$, across a representative calibration dataset. The expectation is approximated as the average across all samples in the calibration set. The search is carried out by grid scanning over a bounded integer parameter space. Although the deployment target may be int8 output, we suggest minimizing the KL-divergence in the int16 space (i.e. comparing against the int16 normalized probabilities instead of the uint8 outputs). Minimizing the int8 KL-divergence produces slightly poorer int8-output KL-divergence in practice, due to the presence of local optima in the quantization rounding at 8-bit precision. The int16 probability representation yields a smoother objective function, and the optimized parameters generally have good transfer performance to the uint8 output path. Finally, for completeness, Algorithm~\ref{alg:hccs} outlines the full inference-time computation of HCCS for a single row.

\begin{algorithm}[t]
\caption{HCCS Softmax Proxy — Inference (Single Row)}
\label{alg:hccs}
\begin{algorithmic}[1]
\STATE \textbf{Input:} Quant.\ logits $\mathbf{x} \in \mathbb{Z}_8^n$,
       $(B_h, S_h, D_{\max,h})$, scale $T$
\STATE \textbf{Output:} Scaled integer probabilities
       $\hat{\mathbf{p}} \in \mathbb{Z}^n$
\STATE $m \leftarrow \max_i x_i$
\FOR{$i = 1$ to $n$}
    \STATE $\delta_i \leftarrow \min(m - x_i,\; D_{\max,h})$
    \STATE $s_i \leftarrow B_h - S_h \cdot \delta_i$
\ENDFOR
\STATE $Z \leftarrow \sum_{i=1}^{n} s_i$
\STATE $\rho \leftarrow \lfloor T / Z \rfloor$
\FOR{$i = 1$ to $n$}
    \STATE $\hat p_i \leftarrow s_i \cdot \rho$
\ENDFOR
\RETURN $\hat{\mathbf{p}}$
\end{algorithmic}
\end{algorithm}

\section{Hardware Implementation on AMD Versal AI Engine}
\label{sec:aie_impl}

\begin{figure*}[t]
  \centering
  \includegraphics[width=1\textwidth]{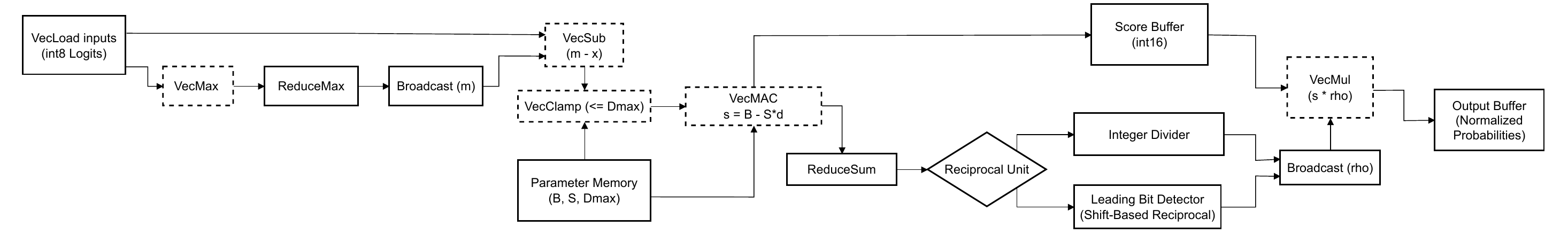}
  \caption{
    Illustration of the softmax proxy pipeline on hardware. Five stages; vector max reduction, unsigned distance and clamp, affine score via int8 MAC, sum reduction, and reciprocal-based normalization are all expressed as integer vector operations with no floating-point or LUT-based exponential at any stage.
  }
  \label{fig:softmax_hw_design}
\end{figure*}

In this section, we describe how the HCCS algorithm can be mapped to an AIE Kernel and how the hardware-based changes were made to create a high-throughput data path using int8. In particular, our design focuses on the common case of attention shape where we apply softmax row-wise over the key dimension (the columns), and process multiple independent rows (queries) in parallel. Furthermore, since processing each row independently, it is natural to partition the row dimension into multiple parallel processing tasks on different AIE Kernels.

\vspace{-0.13cm}
\subsection{Kernel Overview}

For each row of length $n$ the kernel executes five stages (see Figure~\ref{fig:softmax_hw_design}):
\begin{enumerate}
\item \textbf{Vector max reduction:} compute $m = \max_i x_i$ using
      vectorized \texttt{max} operations followed by a horizontal reduce.
\item \textbf{Unsigned distance and clamp:} compute
      $\delta_i = \min(m - x_i,\, D_{\max,h})$ in \texttt{uint8},
      then bit-reinterpret as \texttt{int8} for the MAC stage.
\item \textbf{Affine score via int8 MAC:} compute
      $s_i = B_h - S_h \delta_i$ using a vectorized int8 multiply--accumulate
      into 32-bit accumulators; store $s_i$ in \texttt{int16}.
\item \textbf{Sum reduction:} compute $Z = \sum_i s_i$ accumulated in
      32-bit.
\item \textbf{Reciprocal-based normalization:} compute a fixed-point
      reciprocal of $Z$ (exact division or CLB approximation) and
      multiply each $s_i$ by this reciprocal, emitting normalized
      probabilities in \texttt{uint16} or \texttt{uint8}.
\end{enumerate}

There is one scalar operation, namely the single reciprocal calculation of Z, but the cost of this operation is amortized over the entire row length. Each of the remaining operations are represented as a set of vectorized integer MAC, MUL, and SUB instructions that match the native execution units of the AIE. This contrasts with the BF16 reference design, which uses BF16 exponentials in the AIE-MLv2 architecture or LUT assisted exponentials in the AIE-ML and thus does not operate within the 8 bit pipeline.

\paragraph{Accumulator and normalization overflow analysis.}
Per-element scores satisfy $s_i \le B_h \le \lfloor 32767/n \rfloor$.
The per-lane \texttt{32b} accumulator on AIE gives a headroom of $2^{40} / 32767 \approx 13.1 \times 10^4$ elements before overflow, far beyond any practical attention sequence length. The binding overflow constraints arise in the normalization stage, where the reciprocal $\rho$ is computed as int32 but broadcast into int16 vector lanes. For the int16 output path, the constraint $n \cdot B_h \le 32767$ ensures $Z \le 32767$, so $\rho = \lfloor 32767 / Z \rfloor \ge 1$ without requiring a clamp. For the int8 output path, $\rho = \lfloor 255 \cdot 2^{15} / Z \rfloor \le 32767$ requires $Z \ge 256$, enforced by the calibration floor $B_h - S_h D_{\max,h} \ge \lceil 256/n \rceil$. Together these two constraints define a valid operating band for $B_h$:
\begin{equation}
S_h D_{\max,h} + \left\lceil \frac{256}{n} \right\rceil
\;\le\; B_h
\;\le\; \left\lfloor \frac{32767}{n} \right\rfloor.
\label{eq:Bh_band}
\end{equation}
All intermediate products $s_i \cdot \rho$ are bounded by
$B_h \cdot \lfloor 32767 / (n \cdot B_h) \rfloor \cdot n \le 32767$,
remaining safely within int32 range.

\subsection{Hardware-Aligned Modifications to the Surrogate}
\label{subsec:hw_mods}

In comparison to the surrogate implementation defined in Section \ref{sec:hccs_impl}, the AIE kernel implementation provides the following modifications to increase the overall throughput of the algorithm:

\paragraph{(1) Max subtraction reordered to stay in \texttt{uint8}}

Rather than first calculating the signed centered distance $d_i = x_i - m$ and then negating, we instead calculate  $\delta_i = m - x_i$ directly in unsigned integers (Equation \eqref{eq:uint_distance}). This ensures that the values of $\delta_i$ will always be in the range of $[0, 255]$ and allows us to efficiently perform vector subtraction and clamp in uint8 before casting to int8 for use in the MAC. The transformation is algebraically equivalent to the original transformation and maintains the order of the ranks.

\paragraph{(2) Explicit zero-clamp eliminated by construction}

We established in Section \ref{subsec:int_form} that the requirement to satisfy the calibration constraint  $B_h \ge S_h D_{\max,h}$ will ensure that $s_i \ge 0$ for all clamped distances. Therefore, there is no need for a per-lane  $\max(0,\cdot)$ instruction in the score stage of the design. Thus, we eliminate instructions from the score stage of the design and improve the pipeline efficiency.

\subsection{Integer Range Constraints}
\label{subsec:int_range_constraints}

To guarantee correctness of the int8 MAC stage and the int16 vector normalization pipeline, the following constraints define the admissible parameter region and are enforced during calibration:
\begin{itemize}
\item $D_{\max,h} \le 127$: clamped distances remain representable
      in signed int8.
\item $B_h - S_h D_{\max,h} \ge 0$: all surrogate scores are
      non-negative (Section~\ref{subsec:int_form}).
\item $B_h \le 32767$: int16 storage of scores is safe.
\item $n \cdot (B_h - S_h D_{\max,h}) \ge 256$: ensures the row
      sum $Z \ge 256$, so that the reciprocal $\rho$, computed
      in int32 and broadcast as int16 into the vector normalization
      pipeline, satisfies $\rho \le 32767$ and does not overflow
      int16. This is the binding constraint for the int8 output
      path where $\rho = \lfloor 255 \cdot 2^{R} / Z \rfloor$
      with $R=15$, and it is satisfied by enforcing a minimum
      score floor $B_h - S_h D_{\max,h} \ge \lceil 256/n \rceil$
      during calibration.
\item $n \cdot B_h \le 32767$: ensures the row sum $Z \le 32767$
      so that the reciprocal $\rho = \lfloor 32767 / Z \rfloor \ge 1$
      without requiring a clamp, keeping the normalization valid.
      This bounds the maximum usable $B_h \le \lfloor 32767/n \rfloor$
      and is the tightest upper constraint on $B_h$ for a given
      sequence length $n$.
\end{itemize}

\subsection{Vectorization and Parallelism Across Rows}

Let the softmax input be a 2D tile $\mathbf{X} \in \mathbb{Z}_8^{R \times C}$ where softmax is applied across $C$ columns for each of $R$ independent rows. We will process each row independently with vector width $V$ (for example, $V = 32$) to get $C / V$ vector iterations per row. Since the rows are all independent, the overall processing speed can scale through creating a number of separate AIE kernels. 
\vspace{0.1cm}
\begin{equation}
\{0,\dots,R-1\} = \bigcup_{k=0}^{K-1} \mathcal{R}_k,
\qquad \mathcal{R}_k \cap \mathcal{R}_{k'} = \emptyset,\ k \neq k'.
\end{equation}

Each AIE kernel will compute the softmax for rows in $\mathcal{R}_k$, and this parallelism matches common attention-based workloads that compute many query positions per layer and does not require any synchronization among the kernels other than concatenating results after each has finished.

\section{Evaluation}
\label{sec:eval}

We evaluate HCCS along two orthogonal axes: (i) numerical fidelity relative to standard floating-point softmax, and (ii) throughput on the AMD Versal AI Engine platform.
\vspace{-0.1cm}
\subsection{Experimental Setup}
\label{subsec:exp_setup}

\paragraph{Hardware}
All measurements were obtained using the cycle-accurate AIE simulator provided in the AMD Vitis~2025.2 toolchain which is a standard and widely accepted practice in both FPGA and AIE-related research. AIE benchmarks target the AIE-ML and AIE-MLv2 generation of Versal devices
(VEK280 and VEK385 platforms respectively). Input data is modeled as delivered directly via PLIO, excluding PS/DDR transfer overheads, consistent with streaming inference pipelines.

\paragraph{Baseline}
The baseline was AMD's reference BF16 softmax kernel from IRON API. To the best of our knowledge, it is the only optimized softmax kernel available for this hardware. The BF16 kernel performs an LUT assisted exponential operation on AIE-ML and uses a native BF16 exponential operation instruction on AIE-MLv2. The LUT implementation follows the vendor-provided BF16 exponential primitive, which relies on lookup tables accessed via vector gather instructions and therefore assumes a specific memory layout contract between LUT storage and vector lanes.

\paragraph{Models and tasks}
We evaluate two encoder Transformer models on two classification datasets: binary sentiment classification on SST-2~\cite{Socher2013RecursiveDM} and natural language inference on MNLI~\cite{Williams2017ABC}. The models are BERT-tiny (2 layers, 2 heads, hidden\,=\,128) and BERT-small (4 layers, 8 heads, hidden\,=\,512)~\cite{turc2019wellreadstudentslearnbetter}. The choice of these models represents the lower and upper bounds of what is typically considered to be the low to mid range number of parameters for edge inference on custom hardware like the AMD Versal AI Engine. We intentionally restrict evaluation to encoder-only classification tasks, as these represent the primary deployment target for edge inference on custom integer hardware, and because the softmax overhead is proportionally most significant in this regime relative to GEMM cost. All of the SST-2 experiments utilized a maximum sequence length of 64 tokens. This allows us to cover greater than 99\% of SST-2 training examples without having to truncate them (p99\,=\,44 tokens, max\,=\,66 tokens on the tokenized version of the SST-2 training split). MNLI contains sentence pairs and thus, we had to increase the maximum sequence length to 128 tokens to minimize the number of instances where we would have to truncate the input sequences, while still allowing us to run the experiments at the same throughput levels.

\paragraph{Calibration.}

The calibration process involves determining the set of parameters, $\theta_h$, for each head, $h$. The parameters are $(B_h, S_h, D_{\max,h})$, and are found off-line through the grid search discussed in section~\ref{subsec:calibration_and_frozen} using 64 batch samples. The grid search is designed to find the combination of parameters that results in the minimum average KL-divergence calculated during the grid search in int16 arithmetic. Once a set of parameters is determined, they are used to calibrate the model for the remainder of the experiment. As the calibration process requires no additional computation time beyond finding the optimal parameters, it is performed once per model without introducing additional inference overhead.

\begin{figure*}[!t]
  \centering
  \includegraphics[width=0.95\textwidth]{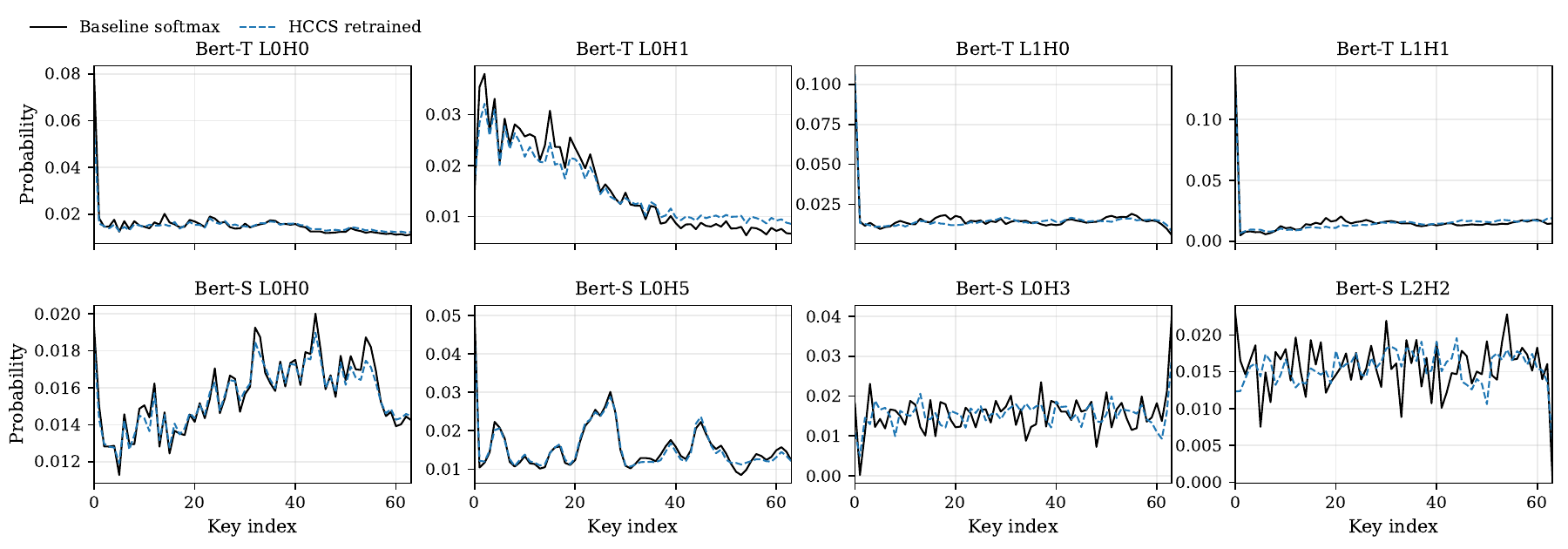}
\caption{Attention probability curves for broad and focused heads from BERT-Tiny (top) and selected BERT-Small heads (bottom), comparing float32 softmax and retrained HCCS. Differences in absolute probabilities are expected, especially on focused heads: HCCS is a surrogate rather than an exact softmax replica, and retraining adapts the model to a new but effective attention distribution.}
  \label{fig:key_probs}
\end{figure*}

\subsection{End-to-End Task Accuracy}
\label{subsec:e2e}

\begin{table}[t]
\centering
\caption{Validation accuracy on SST-2 and MNLI. $\Delta$ is the accuracy
         drop of the retrained HCCS model relative to the
         float32 baseline. Mode: int16+div. MNLI results report matched accuracy on the validation set.}
\label{tab:e2e_accuracy}
\footnotesize
\begin{tabular}{llcccc}
\hline
Task & Model & Baseline & No-retrain & Retrained & $\Delta$ \\
\hline
SST-2 & BERT-tiny  & 0.825 & 0.619 & 0.822 & $-$0.003 \\
SST-2 & BERT-small  & 0.893 & 0.766 & 0.878 & $-$0.015 \\
\hline
MNLI  & BERT-tiny  & 0.653& 0.480 & 0.639 & $-$0.013 \\
MNLI  & BERT-small  & 0.742 & 0.602 & 0.723 & $-$0.019 \\
\hline
\end{tabular}
\end{table}

\begin{table}[t]
\centering
\caption{Effect of lower-granularity calibration after QAT.}
\label{tab:calib_ablation}
\footnotesize
\begin{tabular}{lcc|cc}
\hline
& \multicolumn{2}{c|}{SST-2} & \multicolumn{2}{c}{MNLI} \\
Calibration & BERT-tiny & BERT-small & BERT-tiny & BERT-small \\
\hline
Shared/global & 0.817 & 0.834 & 0.416 & 0.545 \\
Per-layer     & 0.819 & 0.842 & 0.552 & 0.602 \\
\hline
\end{tabular}
\end{table}

To isolate the benefit of calibration granularity, Table~\ref{tab:calib_ablation} reports an ablation study on lower-granularity HCCS parameterizations (shared/global and per-layer) after QAT. In both cases, the proposed head-wise setting (Table~\ref{tab:e2e_accuracy}, ``Retrained'') gives the best downstream accuracy.

The gap is especially pronounced on MNLI, suggesting that heterogeneous heads benefit from finer-grained calibration. Table~\ref{tab:e2e_accuracy} reports validation accuracy for the float32 baseline, direct HCCS substitution without retraining, and HCCS with QAT. After retraining, HCCS remains within 0.3--1.9 percentage points of the float32 baseline across all model-task pairs. These results show that most of the lost accuracy can be recovered with lightweight adaptation of the baseline weights.

The no-retrain results are included to motivate the need for QAT rather than as a practical deployment baseline. The large accuracy drop observed under direct substitution (for example, 20.6 and 12.7 percentage points for BERT-tiny and BERT-small on SST-2, respectively) confirms that retraining is necessary when introducing a surrogate normalization into a compact model. The calibration-stage KL objective is only a proxy for selecting feasible surrogate parameters before retraining; the primary criterion of interest in this work is downstream task accuracy after QAT.

We additionally evaluated the i8+CLB normalization path and observed validation accuracy comparable to the i16+div configuration across all model-task pairs, indicating that the reciprocal approximation has negligible impact on downstream task accuracy in this setting. 

\subsection{Attention Distribution Fidelity}
\label{subsec:fidelity}

A comparison of the attention probability curves in Fig.~\ref{fig:key_probs}, for a representative broad and a focused head in BERT-Tiny, and two selected heads in BERT-Small, using both the float32 baseline and the HCCS retrained models on the same data. Attention entropy was calculated for all heads over multiple samples. Broad heads have the greatest mean attention entropy, while focused Heads have the least. The attention probability curves are plotted against the Key Index (there are 64 keys in both models in SST-2) such that any differences represent the method in which each model allocates its attention over positions. Matching exact probabilities is not required. HCCS is a surrogate softmax, and after retraining the model can converge to different but task-effective attention distributions.

The HCCS model preserves the structural properties of each type of head. 
Broad heads maintain the slow decline of probability over many positions.
Focused heads continue to concentrate their mass into the top ranks. 
However, absolute values of probability differ from the float32 reference, because HCCS is a calibrated monotonic surrogate. In addition, the model will adapt to the HCCS surrogate during training. Therefore, the model is no longer attempting to replicate the float32 softmax, but instead find a different solution in weight space that has a similar loss. The results in the Accuracy table ~\ref{tab:e2e_accuracy} demonstrate that preserving the structural properties of the heads is sufficient for good performance on the target tasks.

The HCCS surrogate has a small KL divergence value compared to the float32 softmax over fixed model weights, as quantified (typically $\approx 0.1$–$0.3$ for broad heads and $\approx 0.2$–$0.3$ for focused heads). The KL divergence values can increase when the model is retrained on the surrogate attention rules, but this did not adversely affect the performance of the model for the downstream tasks (Table~\ref{tab:e2e_accuracy}).



\subsection{Throughput on AMD AI Engine}
\label{subsec:throughput}

Kernel throughput for the BF16 reference and both HCCS configurations are presented for sequence lengths of 32, 64 and 128 in the table~\ref{tab:throughput_samples}, using both AIE-ML and AIE-ML v2. Throughput results represent steady-state kernel execution on the AI Engine. It should be noted that the LUT-based exponential primitive was run as it was delivered within the AMD reference kernel. The actual implementation of this primitive depends on a specific LUT memory layout and gather access contract, thus requiring additional target-specific validation in our Versal AIE-ML implementation. Throughput values of HCCS (CLB) were the largest at all sequence lengths tested. This is attributed to its use of integer MAC operations and a single leading-bit detection instruction instead of the exponential and reciprocal operations used in the reference kernel. The CLB based HCCS kernel eliminates the latency associated with the reciprocal division operation.

\begin{table}[t]
\centering
\caption{Softmax kernel throughput (elements/s) on AMD AI Engine.
         Speedup relative to AMD's BF16 reference per device.}
\label{tab:throughput_samples}
\footnotesize
\begin{tabular}{c|c|cc|cc}
\hline
& BF16
& \multicolumn{2}{c|}{HCCS i16+div}
& \multicolumn{2}{c}{HCCS i8+CLB} \\
$n$ & elems/s
    & elems/s & Speedup
    & elems/s & Speedup \\
\hline
\multicolumn{6}{l}{\textit{AMD Versal VEK280 (AIE-ML)}} \\
\hline
32  & 0.09G/s & 0.41G/s & 4.6$\times$ &1.36G/s & 15.1$\times$ \\
64  & 0.16G/s & 0.78G/s & 4.9$\times$ & 2.19G/s & 13.7$\times$ \\
128 & 0.25G/s & 1.37G/s & 5.5$\times$ & 2.18G/s & 8.72$\times$ \\
\hline
\multicolumn{6}{l}{\textit{AMD Versal VEK385 (AIE-MLv2)}} \\
\hline
32  & 0.24G/s & 0.41G/s & 1.7$\times$ &1.46G/s & 6.1$\times$ \\
64  & 0.46G/s & 0.78G/s & 1.7$\times$ & 2.46G/s & 5.4$\times$ \\
128 & 0.77G/s & 1.41G/s & 1.8$\times$ & 2.21G/s & 2.9$\times$ \\
\hline
\end{tabular}
\end{table}

HCCS (Div), on the other hand, maintains an exact integer division while achieving a throughput that is modestly less than that achieved by the CLB variant. Nevertheless, HCCS (Div) outperformed the BF16 reference kernel at all sequence lengths tested. Throughput generally increased with increasing sequence length due to improved pipeline utilization. At smaller sequence lengths, the overhead of the exponential and reciprocal operations in the BF16 reference kernel was relatively large. In particular, HCCS kernels showed much greater efficiency at lower sequence lengths. As the sequence length increased, the computation became dominated by MAC operations in both kernels. Therefore, the relative throughput improvement decreased somewhat at larger sequence lengths as both kernels approached the limits of the MAC pipelines. Furthermore, under steady-state execution, average row latency increases more slowly than sequence length. For example, in the CLB configuration, it rises from 29 cycles/row at $n=32$ to 69 cycles/row at $n=128$, which is substantially less than a $4\times$ increase. This indicates that fixed per-row costs are amortized over longer rows.

Finally, in the case of the AIE-MLv2, the BF16 baseline had the benefit of a dedicated BF16 exponential instruction. The LUT-based approximation used in the AIE-ML architecture is limited to four parallel table accesses per operation. This limited the throughput of the exponential primitive, while this limitation did not exist in AIE-MLv2 and therefore it achieved a higher throughput on the VEK385 device.

\begin{figure}[!t]
  \centering
  \includegraphics[width=\columnwidth]{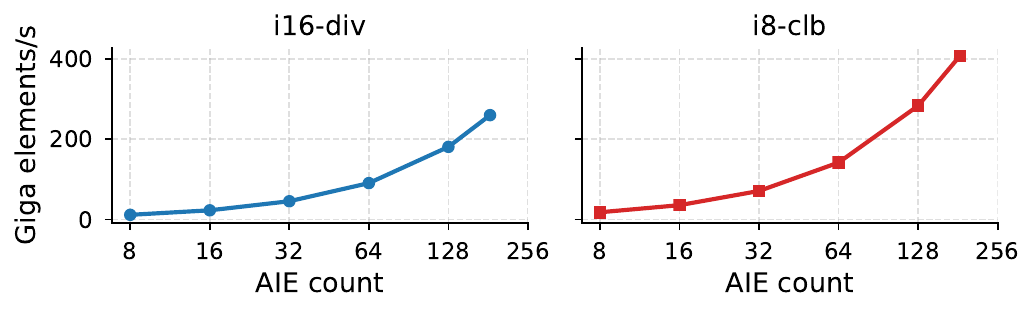}
  \vspace{-0.5cm}
  \caption{Aggregate softmax throughput vs.\ number of AMD
           AI Engine tiles.}
  \label{fig:tile_scaling}
\end{figure}

\paragraph*{Multi-tile scaling}
Figure~\ref{fig:tile_scaling} illustrates aggregate throughput as the number of AI Engine tiles grows from 1 to $184$ on the AIE-MLv2 array (AMD Versal VEK385). The reason we can use multi-tile scaling is that since each softmax row is completely independent, each tile will process a completely different set of rows than all other tiles with no need for inter-tile communication. Each tile also loads the per-head parameters $(B_h, S_h, D_{\max,h})$ for its assigned rows from local tile memory based upon the row's head identifier and therefore does not require any synchronization. Aggregate throughput is obtained by scaling the measured single-tile throughput across independent rows processed on multiple tiles. Due to the embarrassingly parallel nature of the softmax operation, it should be expected that throughput will scale linearly with respect to the number of tiles. Therefore, throughput scales linearly with tile count for both HCCS configurations reaching up to 259~G elements/s for HCCS (i16+Div), and 407~G elements/s for HCCS (i8+CLB) at $184$ tiles. However, as a full DNN workload will not typically allocate such a large portion of the AI Engine array to the softmax stage, this experiment serves to demonstrate the scaling ceiling of the kernel under conditions where enough parallel work is available.

For context, prior FPGA softmax accelerators report operator throughput on the
order of hundreds of M elements/s \cite{mi17010084, app132312784}. Compared to these studies, the maximum throughput of a single AIE tile in the current implementation is approximately an order of magnitude greater than those previously reported. As a result, when scaling across the entire AIE array, the total throughput for the softmax operation will reach hundreds of G elements/s. This level of throughput is similar to that which has been reported for the softmax operation using datacenter-class GPUs, such as the NVIDIA A100 in recent studies \cite{10609604}.

\section{Conclusion}
\label{sec:conclusion}

The paper describes HCCS, which is a head-calibrated clipped-linear softmax surrogate that can be used as an alternative to softmax in transformer attention using an integer-only MAC datapath for edge inference. The per-head parameter calibration of HCCS is completed during off-line training and remains fixed at deployment time, allowing the use of HCCS without changing the surrounding model architecture. Evaluation of the entire pipeline on SST-2 and MNLI with BERT-tiny and BERT-small shows that, with the lightweight retraining needed to recover accuracy loss, we can consistently recover accuracy to less than 2\% of the float32 baseline. Measurements of kernel throughput show that the speedup provided by HCCS (both HCCS i16+Div and HCCS i8+CLB) can be up to $5.5\times$ and up to $15.1\times$, respectively, compared to the vendor BF16 reference on AIE-ML. These results demonstrate that it is possible to achieve both high accuracy and high efficiency when normalizing attention using integer-only methods on production AI Engine hardware.

\section*{Acknowledgment}

 This work has been funded by the Eric \& Wendy Schmidt Fund for Strategic Innovation through the CERN Next Generation Triggers project under grant agreement number SIF-2023-004. M.K. is supported by the US
 Department of Energy (DOE) under Grant No. DE-AC02-76SF00515.

\bibliographystyle{IEEEtran}
\bibliography{my_bib}

\end{document}